\documentclass[journal]{IEEEtran}
\usepackage{caption}
\usepackage{subcaption}
\usepackage{times}
\usepackage{epsfig}
\usepackage{graphicx}
\usepackage{amsmath}
\usepackage{amssymb}
\usepackage{calc}
\usepackage{easybmat}
\usepackage{algpseudocode}
\usepackage{algorithm}
\usepackage{amsthm}
  \newtheoremstyle{dotless}{}{}{\itshape}{}{\bfseries}{}{ }{}
  \theoremstyle{dotless}
  \newtheorem*{thm}{Theorem}

\ifCLASSINFOpdf
\else
\fi
\begin{document}
%
\title{Unseen Face Presentation Attack Detection Using Class-Specific Sparse One-Class Multiple Kernel Fusion Regression}
%
%
%

\author{Shervin~Rahimzadeh Arashloo~\IEEEmembership{}
        
\thanks{S.R. Arashloo is with the department of computer engineering, Faculty of Engineering, Bilkent University, Ankara, 06800, Turkey. E-mail: S.Rahimzadeh@cs.bilkent.edu.tr.}
\thanks{}
\thanks{Manuscript received ? ?, 2020; revised ? ?, 2020.}}

%
%

\markboth{Journal of ?,~Vol.~?, No.~?, ?~?}%
{S.R. Arashloo}
%



\maketitle

\begin{abstract}
The paper addresses face presentation attack detection in the challenging conditions of an unseen attack scenario where the system is exposed to novel presentation attacks that were not present in the training step. For this purpose, a pure one-class face presentation attack detection approach based on kernel regression is developed which only utilises bona fide (genuine) samples for training. In the context of the proposed approach, a number of innovations, including multiple kernel fusion, client-specific modelling, sparse regularisation and probabilistic modelling of score distributions are introduced to improve the efficacy of the method. The results of experimental evaluations conducted on the OULU-NPU, Replay-Mobile, Replay-Attack and MSU-MFSD datasets illustrate that the proposed method compares very favourably with other methods operating in an unseen attack detection scenario while achieving very competitive performance to multi-class methods (benefiting from presentation attack data for training) despite using only bona fide samples for training.
\end{abstract}


%
\IEEEpeerreviewmaketitle

%
%
%
%
\section{Introduction}
\IEEEPARstart{T}{he} functionality of face biometric systems in practical situations is compromised by their susceptibility to presentation attacks (PA's) where an illegitimate trial is made to get access to the system by presenting fake biometric traits to system sensors. Due to potential security risks associated with the problem, face presentation attack detection (a.k.a. anti-spoofing) has received much attention over the past years, resulting in a variety of countermeasures \cite{Bhattacharjee2019,DBLP:series/acvpr/978-3-319-92626-1}. A majority of the existing approaches assume that the face PAD problem is a closed-set recognition task and subsequently formulate and train a binary classifier using the available positive (bona fide) and negative (PA) training samples. Nevertheless, even with the impressive performances reported on some databases, the technology is not mature yet. The discrepancies in samples due to varying image acquisition settings, including sensor interoperability issues, different environmental conditions, etc. degrade the performance of face PAD techniques. In this context, a particularly challenging facet of the problem is due to previously "unseen" attack types for which no similar training samples in terms of the presentation attack instruments are available at training time. In these situations, the common closed-set two-class formulation of the problem tends to generalise poorly, degrading the performance of face PAD methods. While earlier benchmark datasets only included a single attack type (typically printed paper), more recent databases incorporate a more diverse set of attacks, including digital photo attacks, replay attacks, mask attacks, make-up attacks, etc. Although introducing new datasets \cite{8578146,cvpr19yaojie} that cover a wider variety of possible presentation attack mechanisms and instruments is desirable, yet, it may not completely solve the problem. The limitation arises from the fact that not all possible attack scenarios can be anticipated and covered in the datasets since there is always the possibility of developing a new attack strategy to bypass the existing countermeasures. Consequently, error rates obtained on one or more datasets cannot be generalised and regarded as representative error rates corresponding to a real-world operation of the system.

In practice, presentation attacks may potentially appear as fairly diverse, or novel and unpredictable while bona fide sample distributions tend to have relatively less diversity. Notwithstanding other strategies, an alternative approach to the problem is to try to capture the distribution of bona fide samples. This objective may be realised through a one-class classification (OCC) approach to identify patterns from a target class, conforming to a specific condition, and differentiate them from non-target objects. OCC differs from the conventional multi-class formulation in that it mainly relies on samples from a single class for training. In the context of biometric PAD, this approach has been examined, for instance, in speech \cite{voice}, fingerprint \cite{7823572}, or iris anti-spoofing \cite{7760965} as well as in face PAD \cite{7984788,8411206} by considering bona fide samples as target objects and then trying to detect PA's as anomalies w.r.t the population of target observations. The merits of a one-class strategy against the conventional binary-class formulation may be outlined as follows:
\begin{itemize}
\item The multi-modal nature of presentation attack data complicates learning an effective decision boundary as the binary classifier, in this case, tries to associate a potentially very diverse set of attack samples to a single cluster/class. In contrast, since a one-class face PAD method primarily utilises positive (bona fide) samples for training, it would reduce the undesirable impacts of the diversity of presentation attack data on performance. In this respect, the difficulties associated with a sensible choice of the negative (non-target) set common to the conventional binary techniques would be also avoided by virtue of a one-class formulation.
\item In a real-world scenario, attackers may devise inventive ways to fool the system which may not have been available during the training stage of the PAD system. This would in turn have an adverse impact on the performance of the PAD sub-system as the two-class classifiers when exposed to novel (unseen) attack types, tend to exhibit poor generalisation performance. On the contrary, a one-class method is expected to be able to naturally deal with irregular (unknown) patterns by its inherent capability to deal with never seen, non-target observations, and hence, is better equipped to detect unknown presentation attacks.
\item Training a two-class presentation attack detection system requires both genuine as well as attack samples. While collecting genuine access data is relatively easy, increasing the number of presentation attack data for training is more complicated and laborious. Hence, the design of presentation attack detection systems based on a two-class formulation may be affected by a class imbalance problem.
In contrast, in a one-class anomaly detection-based approach it is relatively easier to extend the training set as the decision boundary, in this case, is primarily determined by single-class (i.e. bona fide) observations.
\item The merits of using client-specific solutions over client-independent models in biometrics have been explored in different studies \cite{6905848,Arashloo2017,iet}. However, in the common PAD formulations, the two-class learning scheme inhibits the use of subject-specific solutions in open dataset scenarios, as it is not viable to access subject-specific presentation attack data at the enrolment stage. The one-class formulation, by contrast, is readily amenable to subject-specific model construction in open dataset settings.
\item During the operational course of a biometrics system more samples would become available over time. A common approach in this case is to utilise the thus obtained samples to update the model. While the two-class PAD systems may be not be able to effectively benefit from such data to enhance the performance, a pure anomaly detection-based formulation may readily utilise such data as for training only positive samples are required.
\item And last but not least, from a general pattern classification perspective, since a one-class formulation makes less assumptions regarding the problem at hand, it tends to have a higher capacity to cope with the problems embedded in the nature of the data such as feature and label noise, small sample size, contamination of the data, etc.
\end{itemize}
In practice, based on the availability of training samples, different avenues may be followed to build a one-class classifier. In this study, we conform to the strictest protocol, avoiding the use of PA data altogether (zero-shot \cite{DBLP:journals/corr/abs-1904-02860}) and design the proposed approach in a pure anomaly detection framework utilising only bona fide samples for training.
\subsection{Contributions}
The main contributions of the current work may be summarised as follows:
\begin{enumerate}
\item Following a one-class pure anomaly detection formalism and using only bona fide samples for system training, a new approach to face PAD based on kernel regression is presented. As only bona fide samples are used for training, in the evaluation phase, the detection system is naturally exposed to unseen attacks. 
\item A sparse representation-based variant of the proposed one-class face PAD approach is introduced where for decision making at the operational phase, only a small fraction of the bona fide training samples (as few as 5 frames) may be used, resulting in multiple orders of speed-up gain with respect to its non-sparse counterpart.
\item A client-specific variant of the proposed face PAD approach is introduced by training a separate model for each individual client enrolled in the system. The efficacy of the class-specific solution compared with the client-independent scheme (applying a single global model to detect PA's against all subjects) is validated on different datasets.
\item In the context of the proposed sparse subject-specific one-class model for face PAD, we present a multiple kernel fusion anomaly detection scheme to combine the complementary information provided by different views of the problem for improved detection performance.
\item In terms of the proposed approach, it is illustrated that using generic pre-trained off-the-shelf deep CNN models, it is possible to obtain state-of-the-art performance, even in the challenging conditions of "unseen" presentation attacks.
\item While the proposed approach is applicable to each individual frame in isolation, in order to derive a consensus decision for an entire video sequence, we examine two procedures. The first one applies a mean fusion rule over raw scores corresponding to individual frames. The other assumes a mean fusion over a probabilistic measure of normality. Both alternatives are illustrated to be effective in improving detection performance for a video sequence compared with a single frame.
\item And, lastly, a thorough evaluation of the proposed face PAD approach is carried out on four publicly available datasets in an unseen (zero-shot) presentation attack detection scenario (not utilizing any presentation attack data for training), demonstrating the competitive performance of the proposed approach against other one-class as well as multi-class face PAD techniques.
\end{enumerate}

\subsection{Organisation}
The rest of the paper is organised as follows: Section \ref{RW} reviews related work with an emphasis on unseen face PAD approaches. In Section \ref{BG}, some background material on linear regression and its extension to the reproducing kernel Hilbert space is provided. In Section \ref{Meth}, the proposed one-class approach for unseen face PAD is introduced where sparse regularisation of the solution, client-specific model construction, multiple kernel fusion, frame and video-level decision making are discussed. The results of an experimental evaluation of the proposed approach are presented and discussed in Section \ref{EE}. Finally, conclusions are drawn in Section \ref{conc}.

\section{Related work}
\label{RW}
Different countermeasures including hardware-based, software-based and challenge-response methods \cite{edmunds:tel-01576830} have been proposed for face presentation attack detection. The software-based approaches try to classify an image (sequence) based on different features derived from image content. In this study, we follow a software-based approach to face PAD using a single modality (i.e. visible spectrum RGB images) for PA detection in contrast to some other studies operating beyond the visible spectrum \cite{8794818}. Among the cues conveyed by an image/image sequence for face PAD including texture, motion, frequency, colour, shape or reflectance, texture is the most commonly used feature for face PAD \cite{BOULKENAFET20181,Peng2018}. Deployment  of texture for face PAD is partly fuelled by the assumption that face PA's introduce certain texture patterns that do not occur in bona fide samples. A different category of face PAD methods constitutes motion-based approaches considering either intra-face variations \cite{KOLLREIDER2009233,FENG2016451} or analysing the consistency of a subject's motion with the environment \cite{7185398}. A alternative category of approaches corresponds to frequency-based methods to discern image artefacts in the Fourier domain either from a single image \cite{6199760} or from an image sequence \cite{7017526,7185398}. Colour characteristics \cite{6976921,7351280} as well as shape information are also employed as different cues to deal with presentation attacks \cite{6612957}. Motivated by the assumption that bona fide and PA attempts behave differently under the same illumination conditions, some other methods \cite{10.1007/978-3-642-15567-3_37} use reflectance for face PAD. Other work \cite{8920060} focuses on using a statistical model of image noise for face PAD. A recent category of approaches corresponds to deep learning methods. Inspired by the successful of such models, and in particular, deep convolutional neural networks (CNN's) in different image analysis applications, different methods have been proposed to utilise the discriminatory information of deep CNN representations for face PAD \cite{DBLP:journals/corr/abs-1806-07492,DBLP:journals/corr/YangLL14,8714076}. 

In terms of the two-class classifiers used for decision making, different alternatives have been examined. These include discriminant classifiers with the Support Vector Machines being the most frequently used approach \cite{Heusch2019,LI2018182}. Other classifiers following a discriminative procedure include the linear discriminant analysis  \cite{6976921,HAMDAN201875}, neural networks \cite{FENG2016451}, convolutional neural networks \cite{DBLP:journals/corr/abs-1806-07492,DBLP:journals/corr/YangLL14}, Bayesian networks \cite{EDMUNDS2018314} as well as Adaboost \cite{7880281}. An alternative category includes regression-based approaches trying to project input features  onto their labels \cite{7041231}. Methods trying to learn a distance metric  \cite{10.1007/978-3-642-21605-3_19} also exist in the literature. Some heuristics have been also examined for classification in face PAD \cite{KOLLREIDER2009233,7056504}. 

In contrary to the common close-set two-class formulation of the problem, there also exists a different category of approaches trying to address the face PAD problem in an unseen attack scenario where in the evaluation phase, the system is exposed to PA's which were not available during the training stage of the system. One main group of these methods formulates the problem as an OCC problem to detect unseen PA's \cite{7984788}. In our earlier study \cite{7984788}, a new evaluation protocol as well as a one-class formulation of the face PAD problem based on the anomaly detection concept was presented where the training data came from the positive class only. The evaluation and comparison of 20 different one-class and two-class methods on different databases illustrated that in the presence of unseen PA's, one-class formalism may be considered as a promising direction for investigation. Other work following a one-class formulation \cite{8411206} considers a Gaussian mixture model (GMM) one-class learner for classification operating on image quality measures. Other work \cite{8698574}, analyses One-Class SVM and AutoEncoder-based classifiers to address unseen face PAD. Both approaches, together with the Local Binary Pattern feature are
evaluated and compared on four face PAD datasets. The study in \cite{8682253} examined a new strategy to utilise the client-specific information to train one-class classifiers using only bona fide data. It was found that one-class classifiers trained with client-specific information appeared to be more robust to unseen PA's compared to client-independent and multi-class methods. In an alternative study \cite{DBLP:journals/corr/abs-1904-02860}, detection of unknown PA's was addressed in a Zero-Shot Face Anti-spoofing (ZSFA) scenario via a deep tree network (DTN) to partition the PA samples into semantic sub-groups in an unsupervised fashion. Upon the arrival of new a sample, whether known or unknown in terms of attack type, DTN routes it to the most similar PA cluster to make a binary decision. The work in \cite{DAD} introduces a method where a deep metric learning model is proposed using a triplet focal loss as regularisation for the so-called metric-softmax approach. The benefits of the proposed deep anomaly detection architecture were demonstrated by introducing a few-shot a posteriori probability estimation mechanism.

For a more detailed review of the face PAD methods one may consult \cite{Bhattacharjee2019,Ramachandra:2017:PAD:3058791.3038924,DBLP:series/acvpr/978-3-319-92626-1}.
\section{Background}
\label{BG}
In this section, a brief background on linear regression and its extension to the reproducing kernel Hilbert space (kernel regression) shall be provided. In Section \ref{Meth}, kernel regression is developed as a tool for face PA anomaly detection.
\subsection{Linear Regression}
Let us assume there exists a set of points $x_1,\dots, x_n$ in a $d$-dimensional space, i.e. $x_i\in \mathbb{R}^d$, collected into the matrix $\mathbf{X}^{n \times d}$ which are to be projected onto the real numbers $y_1, \dots, y_n$, i.e. $y_i\in \mathbb{R}$, collectively represented as vector $\mathbf{y}^{n \times 1}$. A commonly followed approach to the problem above is that of linear least squares regression \cite{Alpaydin14} where one assumes that the responses $y_i$'s are generated through a linear process $g(x_i)=\boldsymbol{\alpha}^\top x_i$ and then tries to estimate $\boldsymbol{\alpha}$ by minimising a sum of squared errors:
\begin{eqnarray}
\boldsymbol{\alpha}^{opt} =\operatorname*{arg\,min}_{\boldsymbol{\alpha}}(\mathbf{X}\boldsymbol{\alpha}-\mathbf{y})^\top(\mathbf{X}\boldsymbol{\alpha}-\mathbf{y})
\end{eqnarray}
The solution to the problem above may be found by taking the derivative of the cost function $(\mathbf{X}\boldsymbol{\alpha}-\mathbf{y})^\top(\mathbf{X}\boldsymbol{\alpha}-\mathbf{y})$ w.r.t. $\boldsymbol{\alpha}$ and setting it to zero which yields
\begin{eqnarray}
\boldsymbol{\alpha}=(\mathbf{X}^\top\mathbf{X})^{-1}\mathbf{X}^\top\mathbf{y}
\end{eqnarray}

Given $\boldsymbol{\alpha}$, function $g(.)$ at an arbitrary point $z$ may then be evaluated as
\begin{eqnarray}
\nonumber g(z)&=&\boldsymbol{\alpha}^\top z=\mathbf{y}^\top \mathbf{X}(\mathbf{X}^\top\mathbf{X})^{-1} z=\mathbf{y}^\top(\mathbf{X}\mathbf{X}^\top)^{-1}\mathbf{X}z\\
&=&[\langle z,x_1\rangle,\dots,\langle z,x_n\rangle](\mathbf{X}\mathbf{X}^\top)^{-1}\mathbf{y}
\label{funcg}
\end{eqnarray}
where $\langle .,.\rangle$ denotes inner (dot) product and $.^\top$ represents the transpose operation.
\subsection{Kernel Regression}
Let $\mathcal{F}$ be a feature space induced by a non-linear mapping $\boldsymbol\phi: \mathbb{R}^d \rightarrow \mathcal{F}$. For a suitably chosen mapping, an inner product $\langle.,.\rangle$ on $\mathcal{F}$ may be represented as $\langle \boldsymbol\phi(x_i),\boldsymbol\phi(x_j)\rangle = \kappa(x_i,x_j)$, where $\kappa(.,.)$ is a positive semi-definite kernel function. In kernel regression, each point $x$ is first projected onto $\phi(x)$ followed by seeking a real-valued function $g(\phi(x))=f(x)$ minimising a sum of squared differences between the expected and the generated responses. Using Eq. \ref{funcg}, the relation for $f(z)$ may be written as
\begin{align}
\nonumber f(z)&=[\langle \phi(z),\phi(x_1)\rangle,\dots,\langle \phi(z),\phi(x_n)\rangle](\mathbf{\phi(X)}\mathbf{\phi(X)}^\top)^{-1}\mathbf{y}\\
&=[\kappa(z,x_1),\dots \kappa(z,x_n)]\mathbf{K}^{-1}\mathbf{y}
\end{align}
\noindent where we have used the notation $\mathbf{K}=\mathbf{\phi(X)}\mathbf{\phi(X)}^\top$ to denote the so-called kernel matrix. Denoting $\boldsymbol{\alpha}=\mathbf{K}^{-1}\mathbf{y}$, function $f(.)$ may be represented as
\begin{eqnarray}
\nonumber f(.)&=&[\kappa(.,x_1),\dots \kappa(.,x_n)]\boldsymbol{\alpha}\\
&=&\sum_{i=1}^n \alpha_i \kappa(.,x_i)
\label{funcf2}
\end{eqnarray}
The full responses on the training set $\mathbf{X}$ may be derived as $f(\mathbf{X})=\mathbf{K}\boldsymbol{\alpha}$ and the corresponding cost function in this case is $\Vert\mathbf{K}\boldsymbol{\alpha}-\mathbf{y}\Vert^2$ where $\Vert.\Vert^2$ denotes $L_2$-norm.

\section{Kernel Regression for One-Class Face PAD}
\label{Meth}
In a one-class classification task, it is desirable to have normal samples forming a compact cluster while being distant from anomalies. In a reproducing kernel Hilbert space (RKHS) and in the absence of outlier training data, in a one-class classification paradigm, it is common practice to consider the origin as an artificial exemplar outlier with respect to the distribution of positive samples \cite{Scholkopf:2001:ESH:1119748.1119749}.

In this work, a projection function (defined in terms of kernel regression) is sought such that it maps bona fide samples onto a compact cluster, distant from a hypothetical non-target observation lying at the origin. This objective may be achieved by setting the responses for all target observations to a common fixed real number, distinct from zero, i.e. $y_i=c$, for all $i$, s.t. $c\neq 0$. In this case, the kernel regression approach would form a compact cluster of target samples as they would be all mapped onto the same point. Note that kernel regression performs an exact interpolation when the parameters characterising the regression (i.e. $\boldsymbol{\alpha}$) can be uniquely determined which is the case when the kernel matrix is positive-definite. Since by assumption $c\neq 0$, the projected normal observations would lie away from the (hypothetical) outlier, i.e. the origin. Without loss of generality one may set $c=1$, as the exact value for $c$ would only act as a scaling factor. Having set the response vector $\mathbf{y}$ to $\mathbf{1}^{n\times 1}$, one may solve for $\boldsymbol{\alpha}$ as $\boldsymbol{\alpha}=\mathbf{K}^{-1}\mathbf{y}=\mathbf{K}^{-1}\mathbf{1}^{n\times 1}$.

The procedure discussed above provides the best separability of normal samples from outliers with respect to the Fisher criterion, stated formally in the following theorem.
\begin{thm}
Assuming the origin as a hypothetical outlier, the kernel regression approach with the responses for all target samples set to a common fixed real number other than zero, yields a zero within-class scatter while providing a positive between-class scatter, and thus corresponds to the optimal kernel Fisher criterion for classification, i.e. kernel Fisher null-space.
\end{thm}
\noindent \textbf{Proof} In a Fisher classifier, the criterion function to be maximised is
\begin{eqnarray}
J=\frac{\mathbf{\varphi^\top s_B \varphi}}{\mathbf{\varphi^\top s_W \varphi}}
\end{eqnarray}
\noindent where $\mathbf{s_B}$ is the between-class scatter matrix, $\mathbf{s_W}$ stands for the within-class scatter matrix and $\mathbf{\varphi}$ denotes one axis of the subspace. For the two-category case, the Fisher criterion may be equivalently written as the ratio of the between-class scatter to the total within-class scatter as \cite{Alpaydin14}
\begin{eqnarray}
J=\frac{s_B}{s_W}=\frac{(m_1-m_2)^2}{s_1^2+s_2^2}
\label{sec}
\end{eqnarray}
where $m_1$ and $m_2$ stand for the means of the two classes while $s_1^2$ and $s_2^2$ represent the corresponding within-class scatters. As noted earlier, the Fisher analysis, originally developed for a binary classification problem, may be applied to the one-class scenario by assuming the origin as an exemplar outlier. In this case, the within-class scatter of the transformed bona fide samples is
\begin{eqnarray}
s_1^2=\sum_{C_1}(y_i-m_1)^2=\mathbf{y}^{\top}\mathbf{y}-\frac{1}{n}\mathbf{y}^{\top}\mathbf{1}^{n\times n}\mathbf{y}
\end{eqnarray}
where $C_1$ denotes the target class while $y_i$ denotes the projection of sample $x_i\in C_1$ onto an optimal feature subspace and $m_1$ stands for the mean of the positive class. $\mathbf{y}$ is a vector collection of the responses for transformed data points of the target class while $n$ denotes the number of bona fide samples. 

 As for the negative (presentation attack) class, it is assumed that only a single hypothetical sample exists at the origin, thus $m_2=0$ and $s_2=0$. The total within-class scatter in this case would then be $s_W=s_1^2+s_2^2=s_1^2$.

Regarding the between-class scatter (numerator of $J$ in Eq. \ref{sec}) we have
\begin{eqnarray}
s_B=(m_1-m_2)^2=(\frac{1}{n}\sum_{i=1}^{n}y_{i}-0)^2=\frac{1}{n^2} \mathbf{y}^{\top}\mathbf{1}^{n\times n}\mathbf{y}
\end{eqnarray}
As it is assumed that all target observations are mapped onto $1$, by substituting $\mathbf{y}=(1,\dots,1)^{\top}$ in the relation for the within-class scatter one obtains
\begin{eqnarray}
{s_W}|_{\mathbf{y}=(1,\dots,1)^{\top}}=\mathbf{y}^\top\mathbf{y}-\frac{1}{n}\mathbf{y}^\top\mathbf{1}^{n\times n}\mathbf{y}=0
\end{eqnarray}
Using $\mathbf{y}=(1,\dots,1)^{\top}$, the between-class scatter may be derived as
\begin{eqnarray}
{s_B}|_{\mathbf{y}=(1,\dots,1)^{\top}}=\frac{1}{n^2} \mathbf{y}^{\top}\mathbf{1}^{n\times n}\mathbf{y}=1
\end{eqnarray}
Thus, the one-class kernel regression when the responses of all target (bona fide) samples are equal and distinct from zero, corresponds to a projection function (i.e. $f(.)$) leading to $s_W=0$ and $s_B=1$, and consequently, represents a kernel Fisher null-space analysis \cite{8099922,6619277}. $\blacksquare$

The foregoing kernel regression-based classification approach may be considered as a \textit{one-class} novelty detection variant of our earlier study \cite{7163625} introduced for face PAD and that of \cite{6905848} previously developed for face verification.
\subsection{Regularisation}
Formulating a one-class classifier based on the kernel regression formalism opens the possibility to regularise the solution. Regularising the solution of a regression problem may be driven by different objectives. The first case where regularisation may be applied is when the number of observations is smaller than the number of variables which makes the least-squares problem ill-posed. In this case, regularisation introduces additional constraints to uniquely specify the solution. The second case is when the model suffers from poor generalisation performance. In these circumstances, regularisation serves to improve the generalisation capability of the model by imposing a limitation on the available function space by introducing a penalty to discourage certain regions of the solution space. In such cases, regularisation corresponds to priors on the solution to maintain a trade-off between data fidelity and some constraint on the solution.

A regularised kernel regression problem may be expressed as finding the vector minimser of the following cost function:
\begin{eqnarray}
P(\boldsymbol{\alpha})=\Vert\mathbf{K}\boldsymbol{\alpha}-\mathbf{y}\Vert^2+\mathcal{R}(\boldsymbol{\alpha})
\end{eqnarray}
where, as before, $\Vert\mathbf{K}\boldsymbol{\alpha}-\mathbf{y}\Vert^2$ measures the closeness of the generated responses to the expected responses $\mathbf{y}$ while $\mathcal{R}(\boldsymbol{\alpha})$ encodes a desired regularisation on the solution $\boldsymbol{\alpha}$.

Regularisation schemes favouring sparseness of the solution to derive the simplest possible explanation of an observation in terms of as few as possible atoms from a dictionary are among the widely applied techniques \cite{4483511}. Sparsity of the solution to a least squares problem may be encouraged via an $L_{1}$-norm, the thus obtained problem being known as a basis pursuit in signal processing and Lasso in statistics. In addition to enhanced generalisation performance, sparse $L_{1}$-norm models lead to scalable algorithms that can be applied to problems with a large number of parameters. In this work, encouraging sparseness of the solution of the one-class kernel regression is realised by prescribing an $L_{1}$-regulariser on $\boldsymbol{\alpha}$, i.e. $\mathcal{R}(\boldsymbol{\alpha})=\delta \sum_{i=1}^n |\alpha_i|$. The objective function for the sparse one-class kernel regression may then be expressed as
\begin{eqnarray}
P(\boldsymbol{\alpha}) = \Vert \mathbf{K}\boldsymbol{\alpha}-\mathbf{y} \Vert^2+\delta \sum_{i=1}^n |\alpha_i|
\label{lasso}
\end{eqnarray}
\noindent where parameter $\delta$ controls the trade-off between data fidelity and sparseness of the solution. By imposing a strong sparseness prior on $\boldsymbol{\alpha}$, each response $y_i$ would be characterised by only a few samples from among the training set, an important implication of which is a reduction in the computational cost of the algorithm in the operational stage.
\subsection{Class-Specific Modelling}
The common methodology in the literature for face PAD is to learn a single classifier applicable to all samples, ignoring the identity information. In other words, the PAD sub-system is typically designed in a \textit{client-independent} fashion. The client-independent design scheme rests on the (implicit) assumption that the class label (bona fide/PA) of an observation is independent of the client identity. The client-independent formalism has been re-evaluated in different studies \cite{7031941,8682253,icb_2019_sf} where it has been observed that utilisation of subject-specific information leads to significant improvements in detection performance. Some other work utilising subject-specific information include \cite{7041231} and \cite{7820995}. More specifically, in a class-specific one-class approach for face PAD, one tries to determine what is anomalous with respect to a specific subject as opposed to the class-independent formulation where the goal is to decide what is an outlier in general, irrespective of any specific identity. Training a client-specific face PAD model, however, necessitates subject identities to be accessible both during the training as well as the operational phase of the system. As discussed in \cite{7031941}, such information is readily available to a face PAD engine as it works alongside a face recognition system. As such, in the training phase, the enrolment data associated with each client may be employed to build a subject-specific PAD model. In the operational phase of a face verification system, a test subject claims an identity, while in an identification scenario, the test sample is compared against multiple models stored in the database, whose identities are already known. In both operational modes, the identity of the target class is known and may be deployed by the PAD sub-system to compare a test observation against the model of a specific client. Our earlier study \cite{8682253}, examining a client-specific one-class approach validated the merits of incorporating subject-specific information for the face PAD problem.

Driven by these observations, the current work advocates a client-specific one-class kernel regression approach to the face PAD problem. More specifically, in the present study, during the training phase, a separate one-class kernel regression classifier is built for each subject using the corresponding bona fide data while in the operational stage, the test sample is matched against a subject-specific model.

\subsection{Multiple Kernel Fusion}
The kernel function plays an important role in kernel-based methods as it specifies the embedding of the data in the feature space. While ideally the embedding is to be learnt directly from training samples, in practice, a relaxed version of the problem is considered by trying to learn an optimal combination of multiple kernels providing different views of the problem at hand. The coefficients characterising the combination may then be learnt using training instances from multiple classes \cite{Yan:2012:NMK:2503308.2188406}.  In a one-class novelty detection approach and in the absence of non-target (PA) training samples (i.e. unseen face PAD), we opt for the average fusion of multiple base kernels. In this context, diverse views of the face PAD problem are constructed following two mechanisms: use of multiple (local) face image regions and deployment of multiple representations derived from these regions, discussed next.

\subsubsection{Multiple Regions}
\begin{figure}
     \centering
     \begin{subfigure}[b]{0.08\textwidth}
         \centering
         \includegraphics[scale=.17]{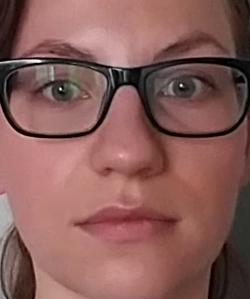}
         \caption{}
         \label{org}
     \end{subfigure}
     \begin{subfigure}[b]{0.08\textwidth}
         \centering
         \includegraphics[scale=.17]{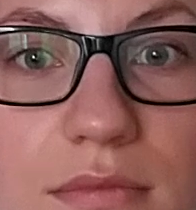}
         \caption{}
         \label{reg1}
     \end{subfigure}
     \begin{subfigure}[b]{0.08\textwidth}
         \centering
         \includegraphics[scale=.17]{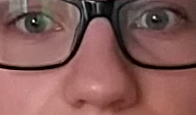}
         \caption{}
         \label{reg2}
     \end{subfigure}
     \begin{subfigure}[b]{0.08\textwidth}
         \centering
         \includegraphics[scale=.17]{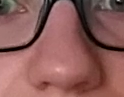}
         \caption{}
         \label{reg3}
     \end{subfigure}
     \begin{subfigure}[b]{0.08\textwidth}
         \centering
         \includegraphics[scale=.17]{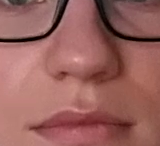}
         \caption{}
         \label{reg4}
     \end{subfigure}
        \caption{Multiple regions used to derive local face representations. (a): image obtained from face detection; (b): region R1; (c): region R2; (d): region R3; (e): region R4.}
        \label{regs}
\end{figure}

In addition to the whole face image providing discriminatory information at a global level, different local regions of the face image convey distinct information for decision making while introducing diversity and providing the possibility to identify facial parts which possess face PA characteristics. In order to benefit from local information, along with the whole face image cropped tightly from face detection bounding box to minimise the background effects (identified as region 1), three additional local regions are also considered. These include eyes and nose as region 2, the areas surrounding the nose as region3 and region 4 which focuses on the areas surrounding the nose and the mouth, Fig. \ref{regs}.

\subsubsection{Multiple Image Representation} The second source of diversity for multiple kernel fusion is derived through using different image representations. While a wide range of different image features including LBP and its variants \cite{George_ICB2019,6905848}, image quality measures \cite{George_ICB2019,8411206}, etc. are applied to  face PAD, more recently, a great deal of research has been directed towards investigating the applicability of deep convolutional neural network (CNN) representations for face PAD \cite{8714076} illustrating that such features may provide discriminative information for detection of presentation attacks. The application of CNN-based representation may also simplify the overall design architecture of biometric systems through adoption of common representation for the face matching problem.  Following the same methodology, the current study utilises CNN representations as features for face PAD. For this purpose, features obtained from the penultimate layers of pre-trained GoogleNet \cite{7298594}, ResNet50 \cite{He2016DeepRL} and VGG16 \cite{Simonyan14c} networks are used to construct multiple representations for each facial region. 
\subsubsection{Learning the discriminant} Once multiple representations from different regions are derived, the learning process of the proposed subject-specific sparse one-class
multiple kernel fusion regression approach for subject $c$ is given as
\begin{eqnarray}
\boldsymbol{\alpha}^c=\operatorname*{arg\,min}_{\boldsymbol{\alpha}}\Vert \frac{1}{RN} (\sum_{r=1}^R\sum_{n=1}^N\mathbf{K}^c_{rn})\boldsymbol{\alpha}-\mathbf{y} \Vert^2+\delta \sum_{i=1}^n |\alpha_i|
\label{sol}
\end{eqnarray}
\noindent where $R=4$ and $N=3$ denote the number of facial regions and deep CNN networks used for face image representation, respectively while $\mathbf{K}^c_{rn}$ stands for the kernel matrix associated with region $r$ whose representation is obtained via CNN with index $n$. In a subject-specific approach, the kernel matrices $\mathbf{K}^c_{rn}$'s are constructed by using the training instances (i.e. features extracted from bona fide frames) of subject $c$ only. In this study, Eq. \ref{sol}, is solved using the Least Angle Regression (LARS) algorithm \cite{efron_2004} which facilitates deriving solutions with all possible cardinalities on $\boldsymbol\alpha$.

\subsection{Decision Strategy}
\label{decr}
Once the projection parameter $\boldsymbol\alpha^c$ is inferred for client $c$, the projection of a test sample ($z$) onto the feature subspace of subject $c$ is given as
\begin{eqnarray}
f^c(z)= \sum_{i=1}^M\alpha_i^c(\frac{1}{RN}\sum_{r=1}^R\sum_{n=1}^N \kappa_{rn}(z,\mathbf{x}_i^c))
\label{proj}
\end{eqnarray}
\noindent where $\alpha_i^c$ denotes the $i^{th}$ (non-zero) element of the discriminant in the Hilbert space for the $c^{th}$ subject while $\mathbf{x}_i^c$ denotes the $i^{th}$ training instance of subject $c$. $M$ is the total number of non-zero elements of $\boldsymbol{\alpha}$. $\kappa_{rn}(z,\mathbf{x}_i^c)$ is the kernel function, capturing the similarity between the $r^{th}$ region of the test sample $z$ and that of the $i^{th}$ training instance of subject $c$ based on the representation derived through the $n^{th}$ deep CNN network.
\subsubsection{Raw score fusion}
Eq. \ref{proj} provides the raw projection score for a single frame of a test video sequence. In order to derive a score for the whole test video, one possibility is to simply average the raw scores corresponding to individual frames comprising the video, leading to a decision rule as
\begin{eqnarray}
\frac{1}{F}\sum_{f=1}^{F}f^c(z_f) \geq \tau^c& \hspace{1cm} &\mbox{\textit{bona fide}}\nonumber \\ 
\nonumber \frac{1}{F}\sum_{f=1}^{F}f^c(z_f) < \tau^c& \hspace{1cm} &\mbox{\textit{PA}}\\
\end{eqnarray}
where $F$ denotes the total number of frames in a video sequence while $\tau^c$ is the threshold for decision making for subject $c$.
\subsubsection{Fusion of probabilistic scores}
Next, we approximate the probability density function of the score distributions corresponding to bona fide samples using a Gaussian function and use the cumulative density function of the inferred Gaussian as a probabilistic measure of normality. In this case, the decision rule for a video sequence reads
\begin{eqnarray}
\frac{1}{F}\sum_{f=1}^{F}\int_{-\infty}^{z_f}\mathcal{N_{\mu,\sigma}}(z_f) \geq \tau^c& \hspace{1cm} &\mbox{\textit{bona fide}}\nonumber \\ 
\nonumber \frac{1}{F}\sum_{f=1}^{F}\int_{-\infty}^{z_f}\mathcal{N}_{\mu,\sigma}(z_f) < \tau^c& \hspace{1cm} &\mbox{\textit{PA}}\\
\label{dr}
\end{eqnarray}
\noindent where $\mathcal{N}_{\mu,\sigma}$ denotes a normal distribution with mean $\mu$ and standard deviation $\sigma$.

\section{Experimental evaluation}
\label{EE}
In this section, the results of an experimental evaluation of the proposed approach on four publicly available datasets in an "unseen" attack scenario are presented. Once the implementation details and the performance metrics are discussed, the requirements for a dataset to be used in a client-specific modelling are outlined. The discussion is then followed by a brief description of the datasets employed in this study. Next, the effects of multiple kernel fusion, client-specific modelling, sparse regularisation and temporal aggregation are presented. Finally, the proposed approach is compared with the state-of-the-art methods in different scenarios. While the proposed approach operates in a zero-shot (unseen attack) scenario, we provide comparisons against other methods which operate in an unseen attack scenario (where no attack data is used for training) as well as multi-class approaches  where both bona fide and attack samples are utilise for training.

\subsection{Implementation Details}
A number of details regarding implementation of the proposed approach are in order. Each frame is initially pre-processed using the photometric normalisation method of \cite{IGI2011} to compensate for illumination variations. The face detection bounding boxes provided along with each dataset are used to locate the face in each frame where in the case of a missing bounding box for a frame, the coordinates of the last detected face in the corresponding video sequence is used. In order to select facial regions (Fig. \ref{regs}) in a consistent fashion across different frames, the OpenFace library \cite{amos2016openface} is used to detect landmarks around facial features. For the extraction of deep CNN representations, their implementations in \textsc{Matlab} are used, yielding 1024-, 2048- and 4096-dimensional feature vectors for GoogLenet (N1), ResNet50 (N2) and VGG16 (N3), respectively. The thus obtained feature vectors are normalised to have a unit $L_2$-norm. The kernel function used is that of a Gaussian (i.e. $\kappa(x_i,z)=\exp({-\theta\Vert x_i-z\Vert^2}$) kernel yielding a positive definite kernel matrix where $\theta$ is set to the reciprocal of the average Euclidean distance between bona fide training samples.
\subsection{Database Requirements for Client-Specific Modelling}
While different datasets are available for the evaluation of face PAD methods, we have chosen four datasets for evaluation of the proposed approach in an unseen PAD scenario. The criteria for this selection not only include presence of different PA instruments, different illumination conditions, different acquisition devices, but also: 1) availability of data for subject-specific model construction, as well as 2) availability of a relatively large number of evaluation results of other face PAD techniques in an unseen scenario for comparison. Regarding the first requirement, one should note that the majority of the existing databases lack a dedicated separate enrolment set of data for subject-specific training. This requirement rules out the possibility to use some data sets including the SiW, SiW-M and other datasets \cite{8578146,10.1007/978-3-642-15567-3_37,cvpr19yaojie} in the context of a client-specific modelling approach. The few exceptions are the Replay-Attack \cite{6313548}, Replay-Mobile \cite{Costa-Pazo_BIOSIG2016_2016} and OULU-NPU datasets \cite{7961798}. While in the Replay-Attack and Replay-Attack datasets a separate enrolment set is provided, the most challenging protocol of the OULU-NPU database (protocol 4) is defined in a way that allows for client-specific modelling. Considering the second criterion for database selection, one notices that a relatively large number of unseen face PAD evaluation results are available on the MSU-MFSD dataset \cite{7031384}.
\subsection{Datasets}
The datasets used in the current work are briefly introduced next.
\subsubsection{The Replay-Mobile dataset}
The Replay-Mobile dataset \cite{Costa-Pazo_BIOSIG2016_2016} includes 1190 video recordings of both bona fide and attack samples of 40 individuals recorded under different illumination conditions using two different acquisition devices. Five different lighting conditions are considered for recording bona fide access attempts. In order to produce the presentation attacks, high-resolution photos and videos from each subjects are captured under two different illumination conditions. The dataset is divided into three disjoint subsets of training, development and testing and an additional partition corresponding to enrolment data. The training set contains 120 bona fide and 192 attack samples while the development set incorporates 160 real-accesses and 256 attacks. The test set is comprised of 110 bona fide videos and 192 presentation attacks. The enrolment set includes 160 bona fide video recordings.

\subsubsection{The Replay-Attack dataset}
The Replay-Attack database \cite{6313548} includes 1300 video recordings of bona fide and attack samples of 50 different individuals. The videos are captured under two different illumination and background conditions. Attacks are created either using a printed image, a mobile phone or a high definition iPad screen. The video recordings in this database are divided randomly into three subject-disjoint subsets for training (60 bona fide and 300 PA samples), development (60 bona fide and 300 PA samples) and testing (80 bona fide and 400 PA samples) purposes. In addition to the training, development and test sets, the Replay-Attack database comes with an extra set of videos recorded in a separate session to serve as enrolment data for each of the 50 clients.

\subsubsection{The OULU-NPU dataset}
The OULU-NPU database \cite{7961798} consists of 4950 bona fide and attack video recordings of 55 subjects recorded using six different cameras in three sessions with different illumination conditions and backgrounds. The dataset includes previously unseen input sensors, attack types and acquisition conditions. The videos of the 55 subjects in this database are divided into three subject-disjoint subsets for training (360 bona fide and 1440 PA samples), development (270 bona fide and 1080 PA samples) and testing (360 bona fide and 1440 PA samples) purposes. For the evaluation purposes, four protocols are designed amongst which, the forth protocol is the most challenging one as in this protocol the evaluation is performed against previously unseen illumination conditions and background scenes, attacks and input sensors. As the test set of the forth evaluation protocol is comprised of samples corresponding to the third session, in order to build client-specific models, the videos from other two sessions may be used.

\subsubsection{The MSU-MFSD dataset}
The MSU MFSD database \cite{7031384} includes 440 videos of photo and video attack attempts of 55 individuals recorded using two different cameras. The publicly available subset, however, includes 35 subjects. The dataset is divided into two subject-disjoint sets for training (30 bona fide and 90 PA's) and testing (40 bona fide and 120 PA's) purposes. While for the previous three datasets there exists independent bona fide data for subject-specific model construction, in this dataset, no such data exists. Thus, for the construction of class-specific models on this dataset, we use $5\%$ of the bona fide data (roughly corresponding to 26 frames per subject) for training and the remaining $95\%$ for testing purposes.

\subsection{Performance Metrics}
For performance reporting ISO metrics BSISO-IEC30107-3-2017 are used in this study \cite{BSISO-IEC30107-3-2017}. These are: 1) attack presentation classification error rate (APCER) which represents the proportion of attack presentations using the same PA instrument species incorrectly classified as bona fide presentations; and, 2) bona fide presentation classification error rate (BPCER) which corresponds to proportion of bona fide presentations incorrectly classified as PA's. When considering how well a PAD subsystem performs in detecting PA's, the APCER of the most successful PA instrument species (PAIS) is reported:
\begin{eqnarray}
APCER=\max_{PAIS}(APCER_{PAIS})
\end{eqnarray}
More specifically, the APCER is computed separately for each presentation attack instrument (e.g. display or print) and the overall presentation attack detection performance corresponds to the most successful attack, i.e. the attack with the highest APCER.

In order to summarised the overall performance of the PAD subsystem as a single number, the Average Classification Error Rate (ACER) defined as the average of the APCER and the BPCER at a specific decision threshold (the threshold where the false rejection rate (FRR) and the false acceptance rate (FAR) are balanced) may be used:
\begin{eqnarray}
ACER = \frac{\max_{PAIS}(APCER_{PAIS}) + BPCER}{2}
\end{eqnarray}
The ACER measure has been deprecated and would be used in here for the sake of completeness and in order to enable a comparison between our results to some other approaches.

In addition to the ISO metrics, the performance of the proposed approach is also reported in terms of the Area Under the ROC Curve (AUC) as well as the Half Total Error Rate (HTER) and the Equal Error Rate (EER), whenever required for a comparison to other techniques.

\subsection{Effect of Multiple Kernel Fusion}
First, the effect of fusing multiple views of the problem via a kernel fusion strategy is analysed. For this purpose and in order to summarise the performance of each individual kernel to facilitate a comparison, ACER's corresponding to different facial regions and different CNN representations are reported in Table \ref{KF} where RxNy denotes the representation for region $x\in\{1,\dots,4\}$ derived from deep CNN $y\in\{\mbox{GoogLenet, ResNet50, VGG16}\}$. From the table, the following observations may be made. First, the best performing regional representation among others in terms of average ACER is R4N3 (i.e. VGG16 applied to the region focusing on the areas surrounding the nose and the mouth). Interestingly, the performance of R4 is better than the whole face image (R1) using the same deep CNN representation. Second, among three different deep representations one observes that the VGG16 network provides the most discriminative features for presentation attack detection. Third, regardless of the region and the deep CNN features employed, the Replay-Mobile and the OULU-NPU (forth protocol) appear to be more challenging as compared with other datasets. In terms of the kernel fusion strategy, the average ACER for the fusion is $5.58\%$ whereas the best performing single kernel provides an average ACER of $7.01\%$ while the worst performing kernel yields an average ACER of $12.83\%$. That is, the kernel fusion improves the average ACER w.r.t. the best single kernel by more than $20\%$ and by more than $56\%$ w.r.t. the worst performing single kernel.

\begin{table}
\renewcommand{\arraystretch}{1.2}
\caption{Effect of Multiple Kernel Fusion in terms of ACER ($\%$). R.M.:Replay-Mobile; R.A.: Replay-Attack; M.M.: MSU-MFSD; O.N.: OULU-NPU; All: average over all datasets.}
\label{KF}
\centering
\begin{tabular}{lccccc}
\hline
\textbf{Kernel} & \textbf{R.M.} & \textbf{R.A.} & \textbf{M.M.} & \textbf{O.N.}& \textbf{All}\\
\hline
R1N1	&20.00& 3.33& 0.04& 13.75& 9.28\\
R2N1	&15.23& 3.12& 0.04& 10.42& 7.20\\
R3N1	&18.18& 8.13& 0.04& 7.50& 8.46 \\
R4N1	&18.41& 8.12& 0.04& 9.17& 8.93\\
R1N2	&30.45& 1.67& 0.04& 11.25& 10.85\\
R2N2	&25.68& 0.21& 0.04& 25.42& 12.83\\
R3N2    &18.64& 0.21& 0.04& 20.42& 9.82\\
R4N2	&19.09& 0&    0.04& 12.08& 7.80\\
R1N3	&20.45& 0.42& 0.04& 10.42& 7.83\\
R2N3	&24.55& 0&    0.04& 8.33& 8.23\\
R3N3    &18.64& 0.42& 0.04& 11.67& 7.69\\
R4N3    &19.90& 0.63& 0.04& 7.50& 7.01\\
\hline
Fusion &15.68& 0& 0& 6.67& 5.58\\
\hline
\end{tabular}
\end{table}
\subsection{Effect of Client-Specific Modelling}
In this section, the efficacy of subject-specific modelling compared to the subject-independent scheme is investigated. In the client-specific approach, a separate one-class kernel regression classifier is built for each subject and subsequently used for decision making. In contrast, in the class-independent approach, a single one-class classifier applicable to all subjects enrolled into the dataset is trained. The performances of the two different approaches in terms of ACER are reported in Table \ref{CS}. From the table, it may be observed that on all four datasets, the client-specific modelling improves the performance. While on the Replay-Attack and MSU-MFSD datasets the client-specific modelling leads to perfect detection performance (zero ACER's), on the Replay-Mobile dataset, it leads to more than $36\%$ improvement in terms of the ACER. The reduction in the error rate on the OULU-NPU dataset is even larger; While the client-independent approach yields an ACER of $43.33\%$ on this dataset, the subject-specific methodology provides an ACER of $6.67\%$. That is, a tremendous improvement of more than $84\%$ is achieved. In terms of the average ACER on the four datasets, the subject-independent approach yields an ACER of $19.11\%$ while the ACER corresponding to the class-specific approach is $5.58\%$, amounting to $70\%$ improvement in ACER, on average.

\begin{table}
\renewcommand{\arraystretch}{1.2}
\caption{Effect of Client-Specific Modelling in terms of ACER ($\%$). R.M.:Replay-Mobile; R.A.: Replay-Attack; M.M.: MSU-MFSD; O.N.: OULU-NPU; All: average over all datasets.}
\label{CS}
\centering
\begin{tabular}{lccccc}
\hline
\textbf{} & \textbf{R.M.} & \textbf{R.A.} & \textbf{M.M.} & \textbf{O.N.}& \textbf{All}\\
\hline
Client-Independent	&24.77& 4.60& 3.75& 43.33& 19.11\\
Client-Specific	&15.68& 0.0& 0.0& 6.67& 5.58\\
\hline
\end{tabular}
\end{table}

\subsection{Effect of Sparse Regularisation}
Next, we examine the effectiveness of sparse regularisation on the one-class kernel regression approach. For this purpose, using the LARS algorithm \cite{efron_2004}, solutions ($\boldsymbol{\alpha}$) with different cardinalities (number of non-zero elements) of 2, 3, 4, 5, 10, 20, 30 and 50 are obtained. Cardinalities of higher than 50 are found not to improve the overall average ACER. The performances corresponding to different cases in terms of ACER are reported in Table \ref{S}. The last column of the table (ARSG), reports the Average Relative Speed-up Gain achieved compared to the non-sparse solution in the test phase. From the table, it may be observed that, interestingly, even by using 2 training frames, one may achieve an impressive overall average ACER of $8.10\%$. Increasing the cardinality from 2 towards 5, a reduction in the average ACER over four datasets is achieved where by using only 5 bona fide training frames, the proposed approach yields an average ACER of $5.58\%$. Increasing the cardinality beyond 5 towards 50, no further improvement in terms of the ACER is obtained. In terms of ARSG, it may be observed that the best performing method in terms of average ACER with NNZ=5 (NNZ: Number of Non-Zero elements in $\boldsymbol\alpha$), yields an impressive 160$\times$ speed-up gain compared to the non-sparse solution.

\begin{table}
\renewcommand{\arraystretch}{1.2}
\caption{Effect of Sparsity in terms of ACER ($\%$). NNZ: number of non-zero elements of $\boldsymbol\alpha$ ; R.M.: Replay-Mobile; R.A.: Replay-Attack; M.M.: MSU-MFSD; O.N.: OULU-NPU; All: average over all datasets; ARSG: average relative speed up gain.}
\label{S}
\centering
\begin{tabular}{lcccccc}
\hline
\textbf{NNZ} & \textbf{R.M.} & \textbf{R.A.} & \textbf{M.M.} & \textbf{O.N.}& \textbf{All}& \textbf{ARSG}\\
\hline
2	&21.36& 0.0& 5.63& 5.42& 8.10& $\approx$ 390$\times$\\
3	&18.86& 0.0& 0.0& 5.83& 6.17& $\approx$ 260$\times$\\
4	&17.50& 0.0& 0.0& 6.67& 6.04& $\approx$ 200$\times$\\
5   &15.68& 0.0& 0.0& 6.67& 5.58& $\approx$ 160$\times$\\
10	&16.14& 0.0& 0.0& 7.50& 5.91& $\approx$ 80$\times$\\
20	&15.68& 0.0& 0.0& 9.17& 6.21& $\approx$ 40 $\times$\\
30 &15.23& 0.0& 0.0& 9.17& 6.10 & $\approx$ 25$\times$\\
50	&15.23& 0.0& 0.0& 10.0& 6.30 & $\approx$ 15$\times$\\
\hline
\end{tabular}
\end{table}
\subsection{Effect of Temporal Aggregation}
In this section, the effect of temporal aggregation of frame-level scores to derive a video-level decision using a sum fusion rule over raw scores and over probabilistic scores is analysed. The frame-level and video-level performances of the proposed approach on different datasets along with the average ACER's are reported in Table \ref{TA}. From the table it may be observed that the sum fusion over raw scores (Video-level (raw)) improves the average ACER on four datasets from $8.99\%$ to $5.58\%$. That is, more than $37\%$ reduction in the average ACER. The average fusion rule applied to the probabilistic scores yields an even further reduction in the average ACER, from $5.58\%$ to $4.97\%$, corresponding to more than $10\%$ improvement.

\begin{table}
\renewcommand{\arraystretch}{1.2}
\caption{Effect of Temporal Aggregation in terms of ACER ($\%$). R.M.:Replay-Mobile; R.A.: Replay-Attack; M.M.: MSU-MFSD; O.N.: OULU-NPU; All: average over all datasets.}
\label{TA}
\centering
\begin{tabular}{lccccc}
\hline
\textbf{} & \textbf{R.M.} & \textbf{R.A.} & \textbf{M.M.} & \textbf{O.N.}& \textbf{All}\\
\hline
Frame-Level	&15.92& 1.91& 4.01& 14.14& 8.99\\
Video-Level (raw)	&15.68& 0.0& 0.0& 6.67& 5.58\\
Video-Level (prob.)	&13.64& 0.0& 0.0& 6.25& 4.97\\
\hline
\end{tabular}
\end{table}

\subsection{Comparison to Other Methods}
In this section, the performance of the proposed sparse one-class kernel fusion regression approach (NNZ=5) is compared against other methods on four datasets. Note that the proposed approach does not utilise any samples for training (i.e. operates in a zero-shot attack scenario). Nevertheless, we provide comparisons between the proposed method and both one-class as well as multi-class approaches.

In addition to other existing face PAD techniques in the literature, in order to provide further baseline performances and enable a more critical assessment of the proposed technique, three other kernel-based one-class classifiers are included for comparison. These are Support Vector Data Description (SVDD) \cite{Tax2004}, Kernel Principal Component Analysis (KPCA) \cite{HOFFMANN2007863} and the Gaussian Process (GP) \cite{KEMMLER20133507}. For a fair comparison, kernel matrices are computed and shared between the proposed method and the other three kernel-based one-class approaches.
\subsubsection{OULU-NPU}
The results of a comparison between the proposed approach and some other methods on the fourth protocol of the OULU-NPU dataset are reported in Table \ref{ONC}. From the table, it may be observed that the average ACER of the proposed approach on the OULU-NPU dataset ($6.25\pm 6.85$) is better than other existing methods in spite of the fact that all other methods in Table \ref{ONC} use PA data for training. The second best performing method on this dataset in terms of ACER is FAS-BAS approach \cite{8578146} with an ACER of $9.5 \pm 6.0$. The GRADIANT method \cite{8272758} with an ACER of $10.0 \pm 5.0$ represents the best performing method on the fourth protocol of the OULU-NPU dataset in the "Competition on Generalized Software-based Face Presentation Attack Detection in Mobile Scenarios" \cite{8272758}. The proposed approach also performs better than other kernel-based (SVDD, KPCA and GP) methods.

\begin{table}[t]
\renewcommand{\arraystretch}{1.2}
\caption{Comparison of the performance of the proposed approach to other methods (including \underline{multi-class} methods) on protocol 4 of the OULU-NPU dataset. (SVDD, KPCA and GP benefit from the same multiple kernel fusion strategy as that of this work.)}
\label{ONC}
\centering
\begin{tabular}{lccc}
\hline
\textbf{Method} & \textbf{APCER($\%$)} & \textbf{BPCER$(\%)$} & \textbf{ACER ($\%$)}\\
\hline
Massy HNU \cite{8272758} &35.8$\pm$35.3&8.3$\pm$4.1&22.1$\pm$17.6\\
GRADIANT\cite{8272758} &5.0$\pm$4.5&15.0$\pm$7.1&10.0$\pm$5.0\\
FAS-BAS\cite{8578146} &9.3$\pm$5.6&10.4$\pm$6.0&9.5$\pm$6.0\\
LBP-SVM\cite{George_ICB2019} &41.67$\pm$27.03&55.0$\pm$21.21&48.33$\pm$6.07\\
IQM-SVM\cite{George_ICB2019} &34.17$\pm$25.89&39.17$\pm$23.35&36.67$\pm$12.13\\
DeepPixBiS\cite{George_ICB2019} &36.67$\pm$29.67&13.33$\pm$16.75&25.0$\pm$12.67\\
SVDD &25.0$\pm$17.32&8.33$\pm$6.83&16.67$\pm$10.68\\
KPCA &13.33$\pm$14.72&11.67$\pm$11.25&12.5$\pm$12.94\\
GP &15.83$\pm$16.25&2.5$\pm$4.18&9.17$\pm$8.76\\
This work &11.67$\pm$13.66&8.3$\pm$2.04 & 6.25$\pm$6.85\\
\hline
\end{tabular}
\end{table}

\subsubsection{Replay-Attack}
The results of a comparison between the proposed approach and other methods in an unseen attack detection scenario on the Replay-Attack dataset are presented in Table \ref{RAC1}. As it may be observed from the table, the proposed approach achieves perfect detection performance on this dataset (AUC=$100\%$) while the best unseen face PAD method from the literature obtains a detection performance of $99.8\%$ in terms of AUC. The proposed method also performs better than the SVDD approach while yielding a similar performance as those of KPCA and GP which both benefit from a similar multiple kernel fusion strategy.

Table \ref{RAC2} presents a comparison between the state of the art multi-class methods and the proposed approach on the Replay-Attack dataset in terms of HTER. The proposed approach achieves a zero HTER, outperforming the majority of multi-class methods. The only method among others achieving a zero HTER is that of \cite{7867821}.

\begin{table}[t]
\centering
\caption{Comparison of the performance of the proposed approach to some other methods on the Replay-Attack dataset in an \underline{unseen} attack scenario in terms of AUC ($\%$).(SVDD, KPCA and GP benefit from the same multiple kernel fusion strategy as that of this work.)}
\label{RAC1}
\begin{tabular}{l c}
\hline
\textbf{Method} &  \textbf{AUC} ($\%$)\\ \hline
OCSVM+IMQ \cite{7984788} & 80.76\\
OCSVM+BSIF \cite{7984788} & 81.94\\
NN+LBP \cite{8698574} & 91.26\\
GMM+LBP \cite{8698574} & 90.06 \\
OCSVM+LBP \cite{8698574}& 87.90\\
AE+LBP \cite{8698574} & 86.12\\
DTL \cite{DBLP:journals/corr/abs-1904-02860} & 99.80\\
SVDD & 97.50\\
KPCA & 100\\
GP & 100\\
This work & 100 \\ \hline
\end{tabular}
\end{table}

\begin{table}[t]
\centering
\caption{Comparison of the performance of proposed approach to the state-of-the-art \underline{multi-class} methods on the Replay-Attack dataset in terms of HTER.
}
\label{RAC2}
\begin{tabular}{lc}
\hline
\textbf{Method}&\textbf{HTER}($\%$)\\
\hline
Boulkenafet et al. \cite{7454730} & 2.90 \\
lsCNN \cite{DBLP:journals/corr/abs-1806-07492} & 2.50 \\
lsCNN Traditionally Trained \cite{DBLP:journals/corr/abs-1806-07492} & 1.75 \\
Chingovska et al. \cite{Chingovska_THESIS_2015} & 6.29 \\
Boulkenafet et al. \cite{7351280} & 3.50 \\
LBP + GS-LBP \cite{Peng2018} & 3.13 \\
Patch-based CNN \cite{8272713} & 1.25 \\
Depth-based CNN \cite{8272713} & 0.75 \\
Fusion of the two Patch and Depth CNNs \cite{8272713} & 0.72 \\
Image Quality Assessment \cite{Fourati2019} & 0.03 \\
Deep Learning \cite{7867821} & 0 \\
This work & 0\\ \hline
\end{tabular}
\end{table}

\subsubsection{MSU-MFSD}
Table \ref{MM1} presents a comparison between the proposed approach and other methods operating in an unseen scenario in terms of AUC. The following observations may be made from the table. The purposed method performs better than other existing face PAD methods, achieving a perfect detection performance as compared to an AUC of $93\%$ for the DTL method \cite{DBLP:journals/corr/abs-1904-02860}. Compared with SVDD, KPCA and GP, the proposed solution performs better than SVDD, while KPCA and GP, benefiting from a similar multiple kernel fusion strategy, achieve a similar performance as that of the proposed approach.

The results of a comparison between this work and the state of the art multi-class methods in terms of EER are presented in Table \ref{MM2}. As it may be observed from the table, the proposed approach (achieving a perfect detection performance) outperforms many multi-class methods. Among others, only the RD method \cite{7820995} provides a similar performance.

\begin{table}[t]
\centering
\caption{Comparison of the performance of the proposed approach to some other methods on the MSU-MFSD dataset in an \underline{unseen} attack scenario in terms of AUC. (SVDD, KPCA and GP benefit from the same multiple kernel fusion strategy as that of this work.)}
\label{MM1}
\begin{tabular}{l c}
\hline
\textbf{Method} &  \textbf{AUC}($\%$) \\ \hline
OCSVM+IMQ \cite{7984788} & 67.77\\
OCSVM+BSIF \cite{7984788} & 75.64\\
NN+LBP \cite{8698574} & 81.59\\
GMM+LBP \cite{8698574} & 81.34\\
OCSVM+LBP \cite{8698574}& 84.47\\
AE+LBP \cite{8698574} & 87.63\\
DTL \cite{DBLP:journals/corr/abs-1904-02860} & 93.00\\
SVDD & 97.5\\
KPCA & 100\\
GP & 100\\
This work & 100 \\ \hline
\end{tabular}
\end{table}

\begin{table}[t]
\centering
\caption{Comparison of the performance of proposed approach to the state-of-the-art \underline{multi-class} methods on the MSU-MFSD dataset in terms of EER ($\%$).
}
\label{MM2}
\begin{tabular}{lc}
\hline
\textbf{Method}&\textbf{EER} ($\%$)\\
\hline
Texture analysis \cite{7454730} & 4.9\\
HSV+YCbCr fusion \cite{7748511} & 2.2\\
Multiscale Fusion \cite{7550078} & 6.9\\
IDA \cite{7031384} & 8.5\\
Colour LBP \cite{7351280} & 10.8\\
HRLF \cite{HRLF} & 0.04\\
RD \cite{7820995} & 0.0\\
This work & 0.0\\ \hline
\end{tabular}
\end{table}

\subsubsection{Replay-Mobile}
A similar comparison as those on other datasets is performed on the Replay-Mobile dataset. In an unseen scenario, the results are reported in Table \ref{RM1} and compared against others in terms of HTER (HTER is chosen for comparison as the majority of the existing unseen PAD results on this dataset have reported their performances in terms HTER). From the table, it may be observed that the proposed approach betters better than other methods in an unseen PAD scenario reported in the literature, achieving a HTER of $13.64$ compared to the best performing unseen method of \cite{8682253} with a HTER of $14.34$. The proposed one-class method also performs better than SVDD, KPCA and GP approaches for face PAD in an unseen attack scenario.

A comparison of the proposed approach to the state of the art multi-class methods is made in Table \ref{RM2}. As it may be observed from the table, the proposed method performs better than some other multi-class methods, yet performs inferior compared to some multi-class techniques.

\begin{table}[t]
\centering
\caption{Comparison of the performance of the proposed approach to some other methods on the Replay-Mobile dataset in an \underline{unseen} attack scenario in terms of HTER.(SVDD, KPCA and GP benefit from the same multiple kernel fusion strategy as that of this work.)}
\label{RM1}
\begin{tabular}{l c}
\hline
\textbf{Method} &  \textbf{HTER} ($\%$)\\ \hline
GoogleNet+SVDD \cite{8682253} &14.34 \\
ResNet50+SVDD \cite{8682253} &21.76\\
VGG16+SVDD \cite{8682253} &18.78\\
GoogleNet+MD \cite{8682253} & 13.70\\
ResNet50+MD \cite{8682253} & 21.81\\
VGG16+MD \cite{8682253} & 19.84\\
GoogleNet+GMM \cite{8682253} & 14.21\\
ResNet50+GMM \cite{8682253} & 21.53\\
VGG16+GMM \cite{8682253} & 18.05\\
SVDD & 16.14\\
KPCA & 17.05\\
GP & 16.36\\
This work & 13.64 \\ \hline
\end{tabular}
\end{table}

\begin{table}[t]
\centering
\caption{Comparison of the performance of proposed approach to the state-of-the-art \underline{multi-class} methods on the Replay-Mobile dataset in terms of HTER.
}
\label{RM2}
\begin{tabular}{lc}
\hline
\textbf{Method}&\textbf{HTER}\\
\hline
two-class SVM + LBP \cite{8411206} & 17.2\\
two-class SVM + Motion \cite{8411206} & 10.4\\
two-class SVM + Gabor \cite{8411206} & 9.13\\
two-class SVM + IQM \cite{8411206} & 4.10\\
This work & 13.64\\ \hline
\end{tabular}
\end{table}

 \subsection{Summary of Performance Evaluations}
Based on the evaluations conducted on four datasets, the proposed approach, when compared to other methods operating in an unseen attack scenario (not using PA samples for training) obtains the-state-of-the-art performance. Moreover, the performance of the proposed method is also very competitive to the state-of-the-art multi-class methods. In this context, on three out of four datasets, the proposed approach achieves better or similar performance compared to methods that benefit from PA training samples for training.

\subsection{A Note on Inter-Dataset Evaluation}
As discussed previously, one of the key contributions of the present work is the introduction of class-specific modelling for face PAD in the context of sparse one-class kernel fusion regression. For this purpose, naturally, subject-specific data is required to build client-specific classifiers. However, as there is not overlap between subjects from different datasets, client-specific approach prevents an inter-dataset evaluation. While the cross-dataset evaluation is expected to be more challenging than the intra-dataset evaluation, it should be noted that the current study has addressed a different, and possibly more challenging problem associated with the face PAD problem, i.e. unseen attack detection. In this respect, the difficulties associated with an inter-dataset evaluation attributed to, for instance, different imaging sensors, different lighting conditions, etc. are somewhat included in the experiments conducted on different datasets such as OULU-NPU which naturally incorporate such variations.

\subsection{Observations}
\begin{itemize}
\item In this study, the methodology was confined to utilising only bona fide samples for training to investigate the potential of a \textit{pure} anomaly detection approach for face PAD. Nevertheless, in the presence of attack samples for training, further refinements to a one-class learner through different mechanisms including a feature selection from among deep CNN representations obtained from local facial regions, a supervised multiple kernel learning approach to learn weights associated with different kernels, etc. may be considered.

\item A further observation corresponds to the Gaussianity assumption made regarding bona fide score distributions (Eq. \ref{dr}). In practice, as observed from the results presented earlier, despite being a simplistic assumption, it led to improvements in detection performance. However, a more suitable distribution model such as those corresponding to Extreme Value Theory \cite{Coles2001} may be considered for improved modelling.

\item A different observation corresponds to a general characteristic of one-class learners. In a novelty detection task where no information regarding anomaly samples is accessible, it is challenging to set the decision thresholds. A common practice in this case is to set the decision threshold at a pre-specified confidence level such that a small proportion of bona fide samples is rejected. In the case of availability of a separate development set specific for each subject, one would be able to set the threshold such that a desired trade-off between FAR and FRR is maintained. In the anomaly approach followed in this work, the HTER corresponds to the confidence level point closest to the EER setting. However, setting the decision threshold in an OCC approach to ensure an suitable practical performance remains an open problem, subject to future investigation.

\item Finally, we note that training a one-class learner over deep CNN representations may be viewed as a new CNN model whose classification layers are constructed using a one-class classifier, where for training, all network parameters except the regression layers are freezed which is considered to be a useful mechanism in the presence of limited training samples.
\end{itemize}

\section{Conclusion}
\label{conc}
The paper addressed the face presentation attack detection problem in the challenging conditions of an unseen attack detection setting. For this purpose, a one-class novelty detection approach, based on kernel regression, was presented. Benefiting from generic deep CNN representations, additional mechanisms including a multiple kernel fusion approach, sparse regularisation of the regression solution, client-specific, and probabilistic modelling were introduced to improve the performance of the system. Experimental evaluation of the proposed approach on four publicly available datasets in an unseen face PAD setting illustrated that the proposed method outperforms other methods operating in an unseen scenario while competing closely with the state-of-the-art multi-class methods. In the case of availability of PA training samples, further improvements may be achieved by refining the solution of a one-class learner through different mechanisms including for instance a feature selection procedure, or multiple kernel learning using multi-class training samples, etc. which are left as future directions of investigation.


%



\ifCLASSOPTIONcaptionsoff
  \newpage
\fi



%
\bibliographystyle{IEEEtran}
\bibliography{ref}

\begin{thebibliography}{10}
\providecommand{\url}[1]{#1}
\csname url@samestyle\endcsname
\providecommand{\newblock}{\relax}
\providecommand{\bibinfo}[2]{#2}
\providecommand{\BIBentrySTDinterwordspacing}{\spaceskip=0pt\relax}
\providecommand{\BIBentryALTinterwordstretchfactor}{4}
\providecommand{\BIBentryALTinterwordspacing}{\spaceskip=\fontdimen2\font plus
\BIBentryALTinterwordstretchfactor\fontdimen3\font minus
  \fontdimen4\font\relax}
\providecommand{\BIBforeignlanguage}[2]{{%
\expandafter\ifx\csname l@#1\endcsname\relax
\typeout{** WARNING: IEEEtran.bst: No hyphenation pattern has been}%
\typeout{** loaded for the language `#1'. Using the pattern for}%
\typeout{** the default language instead.}%
\else
\language=\csname l@#1\endcsname
\fi
#2}}
\providecommand{\BIBdecl}{\relax}
\BIBdecl

\bibitem{Bhattacharjee2019}
S.~Bhattacharjee, A.~Mohammadi, A.~Anjos, and S.~Marcel, \emph{Recent Advances
  in Face Presentation Attack Detection}.\hskip 1em plus 0.5em minus
  0.4em\relax Cham: Springer International Publishing, 2019, pp. 207--228.

\bibitem{DBLP:series/acvpr/978-3-319-92626-1}
S.~Marcel, M.~S. Nixon, J.~Fi{\'{e}}rrez, and N.~W.~D. Evans, Eds.,
  \emph{Handbook of Biometric Anti-Spoofing - Presentation Attack Detection,
  Second Edition}, ser. Advances in Computer Vision and Pattern
  Recognition.\hskip 1em plus 0.5em minus 0.4em\relax Springer, 2019.

\bibitem{8578146}
Y.~{Liu}, A.~{Jourabloo}, and X.~{Liu}, ``Learning deep models for face
  anti-spoofing: Binary or auxiliary supervision,'' in \emph{2018 IEEE/CVF
  Conference on Computer Vision and Pattern Recognition}, June 2018, pp.
  389--398.

\bibitem{cvpr19yaojie}
A.~J. X.~L. Yaojie~Liu, Joel~Stehouwer, ``Deep tree learning for zero-shot face
  anti-spoofing,'' in \emph{In Proceeding of IEEE Computer Vision and Pattern
  Recognition (CVPR 2019)}, Long Beach, CA, 2019.

\bibitem{voice}
A.~O. E.~L. J.~Villalba, A.~Miguel, ``Spoofing detection with dnn and one-class
  svm for the asvspoof 2015 challenge,'' in \emph{16th Annual Conference of the
  International Speech Communication Association}, September 2015.

\bibitem{7823572}
Y.~{Ding} and A.~{Ross}, ``An ensemble of one-class svms for fingerprint spoof
  detection across different fabrication materials,'' in \emph{2016 IEEE
  International Workshop on Information Forensics and Security (WIFS)}, Dec
  2016, pp. 1--6.

\bibitem{7760965}
A.~F. {Sequeira}, S.~{Thavalengal}, J.~{Ferryman}, P.~{Corcoran}, and J.~S.
  {Cardoso}, ``A realistic evaluation of iris presentation attack detection,''
  in \emph{2016 39th International Conference on Telecommunications and Signal
  Processing (TSP)}, June 2016, pp. 660--664.

\bibitem{7984788}
S.~Arashloo, J.~Kittler, and W.~Christmas, ``An anomaly detection approach to
  face spoofing detection: A new formulation and evaluation protocol,''
  \emph{IEEE Access}, vol.~5, pp. 13\,868--13\,882, 2017.

\bibitem{8411206}
O.~Nikisins, A.~Mohammadi, A.~Anjos, and S.~Marcel, ``On effectiveness of
  anomaly detection approaches against unseen presentation attacks in face
  anti-spoofing,'' in \emph{2018 International Conference on Biometrics (ICB)},
  Feb 2018, pp. 75--81.

\bibitem{6905848}
S.~R. Arashloo and J.~Kittler, ``Class-specific kernel fusion of multiple
  descriptors for face verification using multiscale binarised statistical
  image features,'' \emph{IEEE Transactions on Information Forensics and
  Security}, vol.~9, no.~12, pp. 2100--2109, Dec 2014.

\bibitem{Arashloo2017}
S.~R. Arashloo, ``Multiscale binarised statistical image features for symmetric
  face matching using multiple descriptor fusion based on class-specific lda,''
  \emph{Pattern Analysis and Applications}, vol.~20, no.~1, pp. 113--126, Feb
  2017.

\bibitem{iet}
A.~Iosifidis and M.~Gabbouj, ``\BIBforeignlanguage{English}{Neural
  class-specific regression for face verification},''
  \emph{\BIBforeignlanguage{English}{IET Biometrics}}, vol.~7, pp. 63--70(7),
  January 2018.

\bibitem{DBLP:journals/corr/abs-1904-02860}
Y.~Liu, J.~Stehouwer, A.~Jourabloo, and X.~Liu, ``Deep tree learning for
  zero-shot face anti-spoofing,'' \emph{CoRR}, vol. abs/1904.02860, 2019.

\bibitem{edmunds:tel-01576830}
T.~Edmunds, ``{Protection of 2D face identification systems against spoofing
  attacks.}'' Theses, {Univ. Grenoble Alpes}, Jan. 2017.

\bibitem{8794818}
R.~{Tolosana}, M.~{Gomez-Barrero}, C.~{Busch}, and J.~{Ortega-Garcia},
  ``Biometric presentation attack detection: Beyond the visible spectrum,''
  \emph{IEEE Transactions on Information Forensics and Security}, vol.~15, pp.
  1261--1275, 2020.

\bibitem{BOULKENAFET20181}
Z.~Boulkenafet, J.~Komulainen, and A.~Hadid, ``On the generalization of color
  texture-based face anti-spoofing,'' \emph{Image and Vision Computing},
  vol.~77, pp. 1 -- 9, 2018.

\bibitem{Peng2018}
F.~Peng, L.~Qin, and M.~Long, ``Face presentation attack detection using guided
  scale texture,'' \emph{Multimedia Tools and Applications}, vol.~77, no.~7,
  pp. 8883--8909, Apr 2018.

\bibitem{KOLLREIDER2009233}
K.~Kollreider, H.~Fronthaler, and J.~Bigun, ``Non-intrusive liveness detection
  by face images,'' \emph{Image and Vision Computing}, vol.~27, no.~3, pp. 233
  -- 244, 2009, special Issue on Multimodal Biometrics.

\bibitem{FENG2016451}
L.~Feng, L.-M. Po, Y.~Li, X.~Xu, F.~Yuan, T.~C.-H. Cheung, and K.-W. Cheung,
  ``Integration of image quality and motion cues for face anti-spoofing: A
  neural network approach,'' \emph{Journal of Visual Communication and Image
  Representation}, vol.~38, pp. 451 -- 460, 2016.

\bibitem{7185398}
A.~Pinto, H.~Pedrini, W.~R. Schwartz, and A.~Rocha, ``Face spoofing detection
  through visual codebooks of spectral temporal cubes,'' \emph{IEEE
  Transactions on Image Processing}, vol.~24, no.~12, pp. 4726--4740, Dec 2015.

\bibitem{6199760}
G.~Kim, S.~Eum, J.~K. Suhr, D.~I. Kim, K.~R. Park, and J.~Kim, ``Face liveness
  detection based on texture and frequency analyses,'' in \emph{2012 5th IAPR
  International Conference on Biometrics (ICB)}, March 2012, pp. 67--72.

\bibitem{7017526}
A.~Pinto, W.~R. Schwartz, H.~Pedrini, and A.~d.~R.~Rocha, ``Using visual
  rhythms for detecting video-based facial spoof attacks,'' \emph{IEEE
  Transactions on Information Forensics and Security}, vol.~10, no.~5, pp.
  1025--1038, May 2015.

\bibitem{6976921}
J.~Galbally and S.~Marcel, ``Face anti-spoofing based on general image quality
  assessment,'' in \emph{2014 22nd International Conference on Pattern
  Recognition}, Aug 2014, pp. 1173--1178.

\bibitem{7351280}
Z.~Boulkenafet, J.~Komulainen, and A.~Hadid, ``Face anti-spoofing based on
  color texture analysis,'' in \emph{2015 IEEE International Conference on
  Image Processing (ICIP)}, Sept 2015, pp. 2636--2640.

\bibitem{6612957}
T.~Wang, J.~Yang, Z.~Lei, S.~Liao, and S.~Z. Li, ``Face liveness detection
  using 3d structure recovered from a single camera,'' in \emph{2013
  International Conference on Biometrics (ICB)}, June 2013, pp. 1--6.

\bibitem{10.1007/978-3-642-15567-3_37}
X.~Tan, Y.~Li, J.~Liu, and L.~Jiang, ``Face liveness detection from a single
  image with sparse low rank bilinear discriminative model,'' in \emph{Computer
  Vision -- ECCV 2010}, K.~Daniilidis, P.~Maragos, and N.~Paragios, Eds.\hskip
  1em plus 0.5em minus 0.4em\relax Berlin, Heidelberg: Springer Berlin
  Heidelberg, 2010, pp. 504--517.

\bibitem{8920060}
H.~P. {Nguyen}, A.~{Delahaies}, F.~{Retraint}, and F.~{Morain-Nicolier}, ``Face
  presentation attack detection based on a statistical model of image noise,''
  \emph{IEEE Access}, vol.~7, pp. 175\,429--175\,442, 2019.

\bibitem{DBLP:journals/corr/abs-1806-07492}
G.~B. de~Souza, J.~P. Papa, and A.~N. Marana, ``On the learning of deep local
  features for robust face spoofing detection,'' \emph{CoRR}, vol.
  abs/1806.07492, 2018.

\bibitem{DBLP:journals/corr/YangLL14}
J.~Yang, Z.~Lei, and S.~Z. Li, ``Learn convolutional neural network for face
  anti-spoofing,'' \emph{CoRR}, vol. abs/1408.5601, 2014.

\bibitem{8714076}
A.~{George}, Z.~{Mostaani}, D.~{Geissenbuhler}, O.~{Nikisins}, A.~{Anjos}, and
  S.~{Marcel}, ``Biometric face presentation attack detection with
  multi-channel convolutional neural network,'' \emph{IEEE Transactions on
  Information Forensics and Security}, vol.~15, pp. 42--55, 2020.

\bibitem{Heusch2019}
G.~Heusch and S.~Marcel, \emph{Remote Blood Pulse Analysis for Face
  Presentation Attack Detection}.\hskip 1em plus 0.5em minus 0.4em\relax Cham:
  Springer International Publishing, 2019, pp. 267--289.

\bibitem{LI2018182}
L.~Li, X.~Feng, Z.~Xia, X.~Jiang, and A.~Hadid, ``Face spoofing detection with
  local binary pattern network,'' \emph{Journal of Visual Communication and
  Image Representation}, vol.~54, pp. 182 -- 192, 2018.

\bibitem{HAMDAN201875}
B.~Hamdan and K.~Mokhtar, ``The detection of spoofing by 3d mask in a 2d
  identity recognition system,'' \emph{Egyptian Informatics Journal}, vol.~19,
  no.~2, pp. 75 -- 82, 2018.

\bibitem{EDMUNDS2018314}
T.~Edmunds and A.~Caplier, ``Motion-based countermeasure against photo and
  video spoofing attacks in face recognition,'' \emph{Journal of Visual
  Communication and Image Representation}, vol.~50, pp. 314 -- 332, 2018.

\bibitem{7880281}
M.~{Killioğlu}, M.~{Taşkiran}, and N.~{Kahraman}, ``Anti-spoofing in face
  recognition with liveness detection using pupil tracking,'' in \emph{2017
  IEEE 15th International Symposium on Applied Machine Intelligence and
  Informatics (SAMI)}, Jan 2017, pp. 000\,087--000\,092.

\bibitem{7041231}
J.~Yang, Z.~Lei, D.~Yi, and S.~Z. Li, ``Person-specific face antispoofing with
  subject domain adaptation,'' \emph{IEEE Transactions on Information Forensics
  and Security}, vol.~10, no.~4, pp. 797--809, April 2015.

\bibitem{10.1007/978-3-642-21605-3_19}
E.~E.~A. Abusham and H.~K. Bashir, ``Face recognition using local graph
  structure (lgs),'' in \emph{Human-Computer Interaction. Interaction
  Techniques and Environments}, J.~A. Jacko, Ed.\hskip 1em plus 0.5em minus
  0.4em\relax Berlin, Heidelberg: Springer Berlin Heidelberg, 2011, pp.
  169--175.

\bibitem{7056504}
D.~C. Garcia and R.~L. de~Queiroz, ``Face-spoofing 2d-detection based on
  moire-pattern analysis,'' \emph{IEEE Transactions on Information Forensics
  and Security}, vol.~10, no.~4, pp. 778--786, April 2015.

\bibitem{8698574}
F.~{Xiong} and W.~{AbdAlmageed}, ``Unknown presentation attack detection with
  face rgb images,'' in \emph{2018 IEEE 9th International Conference on
  Biometrics Theory, Applications and Systems (BTAS)}, Oct 2018, pp. 1--9.

\bibitem{8682253}
S.~{Fatemifar}, S.~R. {Arashloo}, M.~{Awais}, and J.~{Kittler}, ``Spoofing
  attack detection by anomaly detection,'' in \emph{ICASSP 2019 - 2019 IEEE
  International Conference on Acoustics, Speech and Signal Processing
  (ICASSP)}, May 2019, pp. 8464--8468.

\bibitem{DAD}
D.~Pérez-Cabo, D.~Jiménez-Cabello, A.~Costa-Pazo, and R.~López-Sastre,
  ``Deep anomaly detection for generalized face anti-spoofing,'' 04 2019.

\bibitem{Ramachandra:2017:PAD:3058791.3038924}
R.~Ramachandra and C.~Busch, ``Presentation attack detection methods for face
  recognition systems: A comprehensive survey,'' \emph{ACM Comput. Surv.},
  vol.~50, no.~1, pp. 8:1--8:37, Mar. 2017.

\bibitem{Alpaydin14}
E.~Alpaydin, \emph{Introduction to Machine Learning}, 3rd~ed., ser. Adaptive
  Computation and Machine Learning.\hskip 1em plus 0.5em minus 0.4em\relax MIT
  Press, 2014.

\bibitem{Scholkopf:2001:ESH:1119748.1119749}
B.~Sch\"{o}lkopf, J.~C. Platt, J.~C. Shawe-Taylor, A.~J. Smola, and R.~C.
  Williamson, ``Estimating the support of a high-dimensional distribution,''
  \emph{Neural Comput.}, vol.~13, no.~7, pp. 1443--1471, Jul. 2001.

\bibitem{8099922}
J.~Liu, Z.~Lian, Y.~Wang, and J.~Xiao, ``Incremental kernel null space
  discriminant analysis for novelty detection,'' in \emph{2017 IEEE Conference
  on Computer Vision and Pattern Recognition (CVPR)}, July 2017, pp.
  4123--4131.

\bibitem{6619277}
P.~Bodesheim, A.~Freytag, E.~Rodner, M.~Kemmler, and J.~Denzler, ``Kernel null
  space methods for novelty detection,'' in \emph{2013 IEEE Conference on
  Computer Vision and Pattern Recognition}, June 2013, pp. 3374--3381.

\bibitem{7163625}
S.~R. Arashloo, J.~Kittler, and W.~Christmas, ``Face spoofing detection based
  on multiple descriptor fusion using multiscale dynamic binarized statistical
  image features,'' \emph{IEEE Transactions on Information Forensics and
  Security}, vol.~10, no.~11, pp. 2396--2407, Nov 2015.

\bibitem{4483511}
J.~Wright, A.~Y. Yang, A.~Ganesh, S.~S. Sastry, and Y.~Ma, ``Robust face
  recognition via sparse representation,'' \emph{IEEE Transactions on Pattern
  Analysis and Machine Intelligence}, vol.~31, no.~2, pp. 210--227, Feb 2009.

\bibitem{7031941}
I.~Chingovska and A.~R. dos Anjos, ``On the use of client identity information
  for face antispoofing,'' \emph{IEEE Transactions on Information Forensics and
  Security}, vol.~10, no.~4, pp. 787--796, April 2015.

\bibitem{icb_2019_sf}
S.~{Fatemifar}, M.~{Awais}, S.~R. {Arashloo}, and J.~{Kittler}, ``Combining
  multiple one-class classifiers for anomaly based face spoofing attack
  detection,'' in \emph{ICB 2019 - 2019 IEEE International Conference on
  Biometrics (ICB)}, June 2019.

\bibitem{7820995}
T.~Edmunds and A.~Caplier, ``Fake face detection based on radiometric
  distortions,'' in \emph{2016 Sixth International Conference on Image
  Processing Theory, Tools and Applications (IPTA)}, Dec 2016, pp. 1--6.

\bibitem{Yan:2012:NMK:2503308.2188406}
F.~Yan, J.~Kittler, K.~Mikolajczyk, and A.~Tahir, ``Non-sparse multiple kernel
  fisher discriminant analysis,'' \emph{J. Mach. Learn. Res.}, vol.~13, no.~1,
  pp. 607--642, Mar. 2012.

\bibitem{George_ICB2019}
A.~George and S.~Marcel, ``Deep pixel-wise binary supervision for face
  presentation attack detection,'' in \emph{International Conference on
  Biometrics}, 2019.

\bibitem{7298594}
C.~Szegedy, W.~Liu, Y.~Jia, P.~Sermanet, S.~Reed, D.~Anguelov, D.~Erhan,
  V.~Vanhoucke, and A.~Rabinovich, ``Going deeper with convolutions,'' in
  \emph{2015 IEEE Conference on Computer Vision and Pattern Recognition
  (CVPR)}, June 2015, pp. 1--9.

\bibitem{He2016DeepRL}
K.~He, X.~Zhang, S.~Ren, and J.~Sun, ``Deep residual learning for image
  recognition,'' \emph{2016 IEEE Conference on Computer Vision and Pattern
  Recognition (CVPR)}, pp. 770--778, 2016.

\bibitem{Simonyan14c}
K.~Simonyan and A.~Zisserman, ``Very deep convolutional networks for
  large-scale image recognition,'' \emph{CoRR}, vol. abs/1409.1556, 2014.

\bibitem{efron_2004}
B.~Efron, T.~Hastie, I.~Johnstone, and R.~Tibshirani,
  ``\BIBforeignlanguage{English}{Least angle regression},''
  \emph{\BIBforeignlanguage{English}{The Annals of Statistics}}, vol.~32,
  no.~2, pp. 407--451, 2004.

\bibitem{IGI2011}
V.~\v{S}truc and N.~Pave\v{s}i\'{c}, \emph{Photometric normalization techniques
  for illumination invariance}.\hskip 1em plus 0.5em minus 0.4em\relax
  IGI-Global, 2011, pp. 279--300.

\bibitem{amos2016openface}
B.~Amos, B.~Ludwiczuk, and M.~Satyanarayanan, ``Openface: A general-purpose
  face recognition library with mobile applications,'' CMU-CS-16-118, CMU
  School of Computer Science, Tech. Rep., 2016.

\bibitem{6313548}
I.~Chingovska, A.~Anjos, and S.~Marcel, ``On the effectiveness of local binary
  patterns in face anti-spoofing,'' in \emph{2012 BIOSIG - Proceedings of the
  International Conference of Biometrics Special Interest Group (BIOSIG)}, Sept
  2012, pp. 1--7.

\bibitem{Costa-Pazo_BIOSIG2016_2016}
A.~Costa-Pazo, S.~Bhattacharjee, E.~Vazquez-Fernandez, and S.~Marcel, ``The
  replay-mobile face presentation-attack database,'' in \emph{Proceedings of
  the International Conference on Biometrics Special Interests Group (BioSIG)},
  Sep. 2016.

\bibitem{7961798}
Z.~{Boulkenafet}, J.~{Komulainen}, L.~{Li}, X.~{Feng}, and A.~{Hadid},
  ``Oulu-npu: A mobile face presentation attack database with real-world
  variations,'' in \emph{2017 12th IEEE International Conference on Automatic
  Face Gesture Recognition (FG 2017)}, May 2017, pp. 612--618.

\bibitem{7031384}
D.~Wen, H.~Han, and A.~K. Jain, ``Face spoof detection with image distortion
  analysis,'' \emph{IEEE Transactions on Information Forensics and Security},
  vol.~10, no.~4, pp. 746--761, April 2015.

\bibitem{BSISO-IEC30107-3-2017}
``Information technology — biometric presentation attack detection — part
  3: Testing and reporting,'' International Organization for Standardization,
  Standard, 2017.

\bibitem{Tax2004}
D.~M. Tax and R.~P. Duin, ``Support vector data description,'' \emph{Machine
  Learning}, vol.~54, no.~1, pp. 45--66, Jan 2004.

\bibitem{HOFFMANN2007863}
H.~Hoffmann, ``Kernel pca for novelty detection,'' \emph{Pattern Recognition},
  vol.~40, no.~3, pp. 863 -- 874, 2007.

\bibitem{KEMMLER20133507}
M.~Kemmler, E.~Rodner, E.-S. Wacker, and J.~Denzler, ``One-class classification
  with gaussian processes,'' \emph{Pattern Recognition}, vol.~46, no.~12, pp.
  3507 -- 3518, 2013.

\bibitem{8272758}
Z.~{Boulkenafet}, J.~{Komulainen}, Z.~{Akhtar}, A.~{Benlamoudi}, D.~{Samai},
  S.~E. {Bekhouche}, A.~{Ouafi}, F.~{Dornaika}, A.~{Taleb-Ahmed}, L.~{Qin},
  F.~{Peng}, L.~B. {Zhang}, M.~{Long}, S.~{Bhilare}, V.~{Kanhangad},
  A.~{Costa-Pazo}, E.~{Vazquez-Fernandez}, D.~{Perez-Cabo}, J.~J.
  {Moreira-Perez}, D.~{Gonzalez-Jimenez}, A.~{Mohammadi}, S.~{Bhattacharjee},
  S.~{Marcel}, S.~{Volkova}, Y.~{Tang}, N.~{Abe}, L.~{Li}, X.~{Feng}, Z.~{Xia},
  X.~{Jiang}, S.~{Liu}, R.~{Shao}, P.~C. {Yuen}, W.~R. {Almeida}, F.~{Andalo},
  R.~{Padilha}, G.~{Bertocco}, W.~{Dias}, J.~{Wainer}, R.~{Torres}, A.~{Rocha},
  M.~A. {Angeloni}, G.~{Folego}, A.~{Godoy}, and A.~{Hadid}, ``A competition on
  generalized software-based face presentation attack detection in mobile
  scenarios,'' in \emph{2017 IEEE International Joint Conference on Biometrics
  (IJCB)}, Oct 2017, pp. 688--696.

\bibitem{7867821}
I.~{Manjani}, S.~{Tariyal}, M.~{Vatsa}, R.~{Singh}, and A.~{Majumdar},
  ``Detecting silicone mask-based presentation attack via deep dictionary
  learning,'' \emph{IEEE Transactions on Information Forensics and Security},
  vol.~12, no.~7, pp. 1713--1723, July 2017.

\bibitem{7454730}
Z.~Boulkenafet, J.~Komulainen, and A.~Hadid, ``Face spoofing detection using
  colour texture analysis,'' \emph{IEEE Transactions on Information Forensics
  and Security}, vol.~11, no.~8, pp. 1818--1830, Aug 2016.

\bibitem{Chingovska_THESIS_2015}
I.~Chingovska, ``Trustworthy biometric verification under spoofing attacks:
  Application to the face mode,'' Ph.D. dissertation, {\'{E}}cole Polytechnique
  F{\'{e}}d{\'{e}}rale de Lausanne, Nov. 2015, th{\`{e}}se EPFL, no 6879
  (2016).

\bibitem{8272713}
Y.~{Atoum}, Y.~{Liu}, A.~{Jourabloo}, and X.~{Liu}, ``Face anti-spoofing using
  patch and depth-based cnns,'' in \emph{2017 IEEE International Joint
  Conference on Biometrics (IJCB)}, Oct 2017, pp. 319--328.

\bibitem{Fourati2019}
E.~Fourati, W.~Elloumi, and A.~Chetouani, ``Anti-spoofing in face
  recognition-based biometric authentication using image quality assessment,''
  \emph{Multimedia Tools and Applications}, Oct 2019.

\bibitem{7748511}
Z.~{Boulkenafet}, J.~{Komulainen}, and A.~{Hadid}, ``Face antispoofing using
  speeded-up robust features and fisher vector encoding,'' \emph{IEEE Signal
  Processing Letters}, vol.~24, no.~2, pp. 141--145, Feb 2017.

\bibitem{7550078}
Z.~{Boulkenafet}, J.~{Komulainen}, {Xiaoyi Feng}, and A.~{Hadid}, ``Scale space
  texture analysis for face anti-spoofing,'' in \emph{2016 International
  Conference on Biometrics (ICB)}, June 2016, pp. 1--6.

\bibitem{HRLF}
U.~Muhammad and A.~Hadid, ``Face anti-spoofing using hybrid residual learning
  framework,'' in \emph{2019 International Conference on Biometrics (ICB)},
  June 2019.

\bibitem{Coles2001}
S.~Coles, \emph{An introduction to statistical modeling of extreme values},
  ser. Springer Series in Statistics.\hskip 1em plus 0.5em minus 0.4em\relax
  London: Springer-Verlag, 2001.

\end{thebibliography}

%






\end{document}